\documentclass[10pt,twocolumn,letterpaper]{article}

\usepackage[pagenumbers]{wacv} 

\usepackage{graphicx}
\usepackage{amsmath}
\usepackage{amssymb}
\usepackage{booktabs}

\usepackage{kotex}
\usepackage{bbm}
\usepackage{multirow}
\usepackage{tabularx}
\usepackage{array}
\usepackage{hhline}
\usepackage{graphicx}
\usepackage{amsmath}
\usepackage{lipsum}

\usepackage[pagebackref,breaklinks,colorlinks]{hyperref}

\usepackage[capitalize]{cleveref}
\crefname{section}{Sec.}{Secs.}
\Crefname{section}{Section}{Sections}
\Crefname{table}{Table}{Tables}
\crefname{table}{Tab.}{Tabs.}

\def\wacvPaperID{1347} 
\def\confName{WACV}
\def\confYear{2025}

\begin{document}

\title{Semi-Supervised Audio-Visual Video Action Recognition \\ with Audio Source Localization Guided Mixup}


\author{Seokun Kang,\quad Taehwan Kim\\
Artificial Intelligence Graduate School\\
Ulsan National Institute of Science \& Technology\\
{\tt\small \{oraclemiso, taehwankim\}@unist.ac.kr}
}
\maketitle

\newcommand{\mycolor}{black}

\begin{abstract}
Video action recognition is a challenging but important task for understanding and discovering what the video does. However, acquiring annotations for a video is costly, and semi-supervised learning (SSL) has been studied to improve performance even with a small number of labeled data in the task.
Prior studies for semi-supervised video action recognition have mostly focused on using single modality - visuals - but the video is multi-modal, so utilizing both visuals and audio would be desirable and improve performance further, which has not been explored well. Therefore, we propose audio-visual SSL for video action recognition, which uses both visual and audio together, even with quite a few labeled data, which is challenging. In addition, to maximize the information of audio and video, 
we propose a novel audio source localization-guided mixup method that considers inter-modal relations between video and audio modalities. 
In experiments on UCF-51, Kinetics-400, and VGGSound datasets, our model shows the superior performance of the proposed semi-supervised audio-visual action recognition framework and audio source localization-guided mixup.


\end{abstract}    
\section{Introduction}\label{sec:intro}
Video is considered an important resource for deep learning vision research. Video data of different lengths and formats has become more readily available as large-scale video content offerings and user interactions on online platforms increase. This leads to active research on video understanding~\cite{tong2022videomae, li2023videochat, li2023uniformerv2, liu2022tcgl} and among them, video action recognition is one of the challenging tasks~\cite{yang2023aim, chen2022mm, gowda2022learn2augment, lee2023modality, shaikh2023maivar, zhu2024efficient}. 

Unlike image classification, video action recognition is a challenging task that requires understanding both spatial and temporal information. This requires sufficient labeled training data, but labeling on video data is more difficult and time-consuming than images. Nevertheless, unlabeled video data is readily available, and semi-supervised learning (SSL) methods using labeled video and large-scale unlabeled video together are being actively studied~\cite{tong2022semi, xing2023svformer, xiao2022learning, xu2022cross}.

Semi-supervised video action recognition research remains less explored compared to semi-supervised image classification~\cite{sohn2020fixmatch, zhang2021flexmatch, chen2022debiased, wang2022freematch, chen2023boosting}. Recent studies in SSL video recognition have actively explored various approaches beyond just using fixed, flexible, or even not predefined confidence-based thresholds~\cite{sohn2020fixmatch, zhang2021flexmatch, wang2022freematch} as proposed in the SSL image domain. These include leveraging pre-trained networks and large-scale unlabeled datasets~\cite{dave2023timebalance, xing2023svformer, jing2021videossl}. 
In addition, research efforts are increasingly focusing on the utilization of additional modalities such as temporal gradient~\cite{xiao2022learning} and optical flow~\cite{xiong2021multiview} obtained from video data or by introducing auxiliary networks~\cite{xu2022cross}. 

Studies of semi-supervised learning for image and video use random augmentation techniques to generate weak and strong augmented images or videos and train the model through consistency regularization using both augmented~\cite{sohn2020fixmatch, zhang2021flexmatch, wang2022freematch, xing2023svformer, xu2022cross}. 
In this context, weak augmentations typically involve random flips and rotations, while strong augmentations utilize RandAugment~\cite{cubuk2020randaugment} that includes techniques such as CutMix~\cite{yun2019cutmix}, which involves cutting and combining different images, and MixUp~\cite{zhang2017mixup}, which mixes images through interpolation.

However, SVFormer~\cite{xing2023svformer} points out the limitations of mixup and cutmix that do not produce the expected performance in the video domain, and to overcome these limitations, they propose TubeToken Mixup and Temporary Warping augmentation methods. However, there is still a limitation in that they cannot actively utilize all its information because only visual information is considered, not audio information. 
In semi-supervised video action recognition, using visual-audio modality has been under-explored. The only exception is AvCLR~\cite{assefa2023audio}, which employs audio-visual contrastive learning with ResNet architecture. 
However, despite originating from the same video clip, augmentations for video and audio modalities are conducted individually without profoundly considering the inter-modal relation between these modalities.

In this paper, we propose the transformer-based framework for semi-supervised audio-visual action recognition with audio source localization-guided mixup. Specifically, we introduce a novel audio source localization-guided mixup method designed to preserve the inter-modal relations that tend to be overlooked when applying mixup or cutmix to visual and audio modalities. 
In most cases, visual and audio information occur simultaneously and are interrelated in videos. By preserving this inter-relationship within videos, models can better understand and reflect real-world situations.

In addition, we leverage contrastive learning between video and audio modalities in the video clip. Incorporating audio source localization guided mixup with contrastive learning can improve performance substantially over existing state-of-the-art methods. 

In summary, our contributions are as follows:
\begin{itemize}
\vspace{-6pt}
\item We study semi-supervised audio-visual action recognition, which has been under-explored. To the best of our knowledge, we propose, for the first time, a visual-audio semi-supervised video action recognition approach based on the transformer model. This allows for the maximization of the use of visual and audio information inherent in videos.
\vspace{-6pt}
\item We propose a novel audio source localization-guided mixup that, unlike mixup or cutmix, preserves the interrelation between the video and audio modalities.
\vspace{-6pt}
\item In experiments, our proposed model outperforms the existing state-of-the-art model on three different benchmark datasets in even challenging scenarios with only a few labeled samples available. 

\end{itemize}
\vspace{-4pt}

\section{Related Works}\label{sec:related}
\vspace{-4pt}
\subsection{Semi-supervised Image Classification}
Deep learning-based image classification has achieved significant performance improvements by leveraging large-scale annotated datasets~\cite{szegedy2015going, krizhevsky2012imagenet, he2016deep, howard2017mobilenets, dosovitskiy2020image, liu2021swin}. However, annotating requires considerable time and resources. To address this, semi-supervised learning (SSL) methods, which use a few labeled datasets and large-scale unlabeled datasets, have been introduced. These approaches aim to reduce annotating time and resource consumption while improving performance. The most commonly used techniques include consistency learning through data augmentations~\cite{tarvainen2017mean, berthelot2019mixmatch} and the use of pseudo labels~\cite{laine2016temporal, dong2018tri}. Consistency learning focuses on training the model to make consistent predictions about augmented images, while pseudo-labeling uses the model's predictions on unlabeled data as pseudo-labels. 
\textcolor{\mycolor}{
Additionally, various studies use consistency learning and pseudo-labeling together to reduce bias from pseudo-label via a thresholding approach.
}
For example, FixMatch~\cite{sohn2020fixmatch} uses fixed predefined thresholds, FlexMatch~\cite{zhang2021flexmatch} employs flexible thresholds, and FreeMatch~\cite{wang2022freematch} uses thresholds that are not predefined but are also flexible.

\subsection{Semi-supervised Video Recognition}
As the creation and utilization of video data increase across various platforms, research on video understanding is becoming increasingly active~\cite{tong2022videomae, li2023videochat, li2023uniformerv2, liu2022tcgl, dave2023timebalance}. Among them, video recognition~\cite{mao2018hierarchical, bhardwaj2019efficient, li2022mvitv2, liu2022video, chen2022mm, arnab2021vivit, lee2023modality, shaikh2023maivar, zhu2024efficient} is one of the challenging tasks. Similar to image classification, video recognition requires large amounts of annotated data. However, videos have more complex information than images, which needs significant annotation time and resources. To address this challenge, various studies have been conducted to extend semi-supervised learning techniques used in image classification to video recognition~\cite{xu2022cross, assefa2023actor, tong2022semi, wu2023neighbor}.

VideoSSL~\cite{jing2021videossl} extends the existing semi-supervised image classification methods to the video modality. In addition, MvPL~\cite{xiong2021multiview} has introduced a multi-modal approach that leverages optical flows and temporal gradients derived from videos. On the other hand, SVFormer~\cite{xing2023svformer} recognizes the limitation that image-based augmentation methods, such as mixup, do not perform well in the video domain. To overcome this limitation, SVFormer introduced TubeToken Mixup, which mixes the two different videos at the token level feature. 
In SSL video recognition, studies considering video and audio modalities are scarce, with AvCLR~\cite{assefa2023audio} being a notable exception. However, its augmentation methods focus on individual modalities, overlooking the inter-modal relation between video and audio. Therefore, We introduce a novel approach that employs audio source localization-guided mixup to bridge this gap, considering the interrelation between video and audio.

\subsection{Audio Source Localization}
Audio source localization is the field of research that combines visual and audio information to identify the source of audio within the scene~\cite{senocak2018learning, senocak2023sound, um2023audio}. 
Senocak et al.~\cite{senocak2018learning} propose the two-stream network architecture that uses attention mechanisms for visual and audio modalities through unsupervised learning.
Hu et al.~\cite{hu2022mix} introduce multi-source audio-visual sound localization utilizing contrastive learning and cycle consistency using the synthetic mixture of audio from multiple videos. 
EZ-VSL~\cite{mo2022localizing} proposes multiple instance contrastive learning for alignment between audio-visual modalities and the object-guided localization scheme that utilizes an object localization map generated from a pre-trained visual encoder. FNAC~\cite{sun2023learning} proposes False Negatives Suppression that uses intra-modal similarity to identify potential false negatives and minimize their negative impact, and True Negatives Suppression that encourages more discriminatory sound source localization by highlighting true negatives using different localization results.
We propose an augmentation methodology that utilizes audio source localization to consider the inter-modal relation of video and audio modalities.

\begin{figure*}[htb!]
    \centering
    \includegraphics[trim={0cm 0.5cm 0cm 0cm},width=\textwidth, clip]{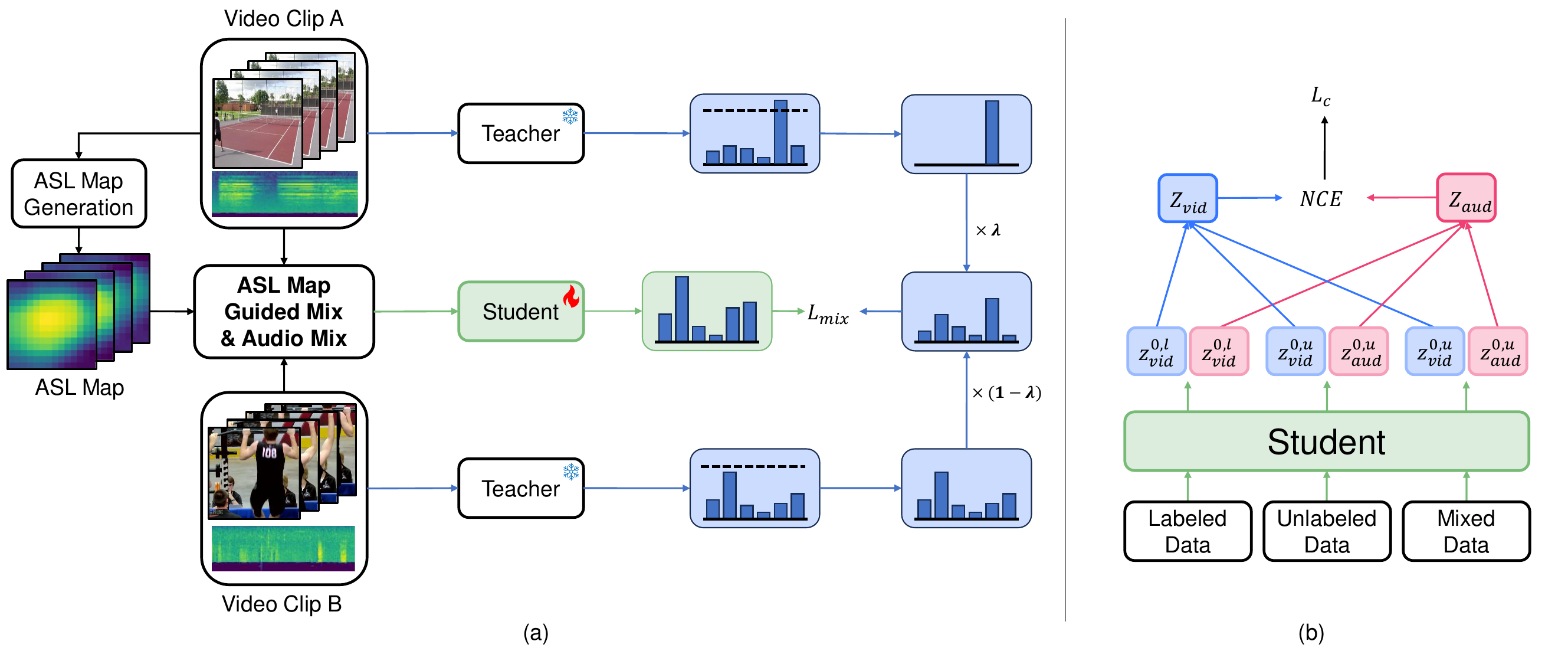}
    \caption{{\bf Overview of audio source localization-guided mixup framework.} The framework performs audio source localization-guided mixup on video clips A and B. A localization map from video A is used to create a mask that highlights semantically important regions, considering the interrelation between video and audio. Log mel-filterbank coefficients of audios A and B are then interpolated, and the mixed video and audio are used for prediction.\vspace{-10pt}
}
    \label{fig:framework}
\end{figure*}

\section{Methodology}\label{sec:method}
In this section, first, we describe the methodologies primarily used in semi-supervised learning. Then, we introduce the multi-modal video recognition framework that extends to video-audio modalities from the existing video recognition framework. We then introduce our proposed audio source localization-guided mixup methodology.

\subsection{Preliminaries}
In semi-supervised learning, there are few labeled samples and large-scale unlabeled samples. Let $D_{L}=\{(x^{l}, y^{l}): l \in [N_{L}]\}$ as labeled dataset and $D_{U}=\{(x^{u}): u \in [N_{U}]\}$ as unlabeled dataset where $N_{L}$ and $N_{U}$ are numbers of samples in labeled and unlabeled dataset respectively, and in general $N_{U} \gg  N_{L}$. For the labeled dataset, like the general classification problem, training is conducted through standard cross-entry loss between predicted probabilities and labels like Eq.~\ref{eq:ce-loss}.

\small
\begin{equation}
    L_{s}=\frac{1}{B_{l}}\sum_{i}^{B_{l}}H(y^{l}_{i}, p_{m}(x^{l}_{i})))
\label{eq:ce-loss}
\end{equation}
\normalsize

where $B_{l}$ is batch size of the labeled data and $p_{m}(\cdot)$ is denoted predicted probability from the model. 
In the case of the unlabeled data $x^u$, consistency regularization is used through data augmentation like weak-augmentation $\alpha_{vid}(\cdot)$~(e.g., random flip or rotation) and strong-augmentation $\boldsymbol{A}_{vid}(\cdot)$~(e.g., 
 RandAugmentation~\cite{cubuk2020randaugment}). Throughout these two different augmentation levels, consistency learning is conducted for unlabeled data like Eq.~\ref{eq:consystency-loss}.

\vspace{-5pt}
\small
\begin{equation}
    L_{u}=\frac{1}{B_{u}}\sum_{i}^{B_{u}}H(\hat{p}_{m}(\alpha_{vid}(x^{u}_{i}))), p_{m}(\mathit{A}_{vid}(x^{u}_{i})))
\label{eq:consystency-loss}
\end{equation}
\normalsize
where $B_u$ is the number of unlabeled data in batch and $\hat{p}_{m}(\cdot)$ can be predicted probability from the Mean Teacher model or model itself by freezing the parameters. And let $\hat{p}_{m}(\alpha_{vid}(x^{u}_{i}))$ as $\hat{q}^{u}_{i}$ and $p_{m}(\mathit{A}_{vid}(x^{u}_{i}))$ as $Q^{u}_{i}$. In this case, the predicted probability served as the pseudo label is not reliable, so FixMatch~\cite{sohn2020fixmatch} proposes using fixed threshold $\tau$ for remaining confidential probability as the pseudo label like Eq.~\ref{eq:fixmatch}. 

\vspace{-5pt}
\small
\begin{equation}
    L_{u}=\frac{1}{B_{u}}\sum_{i}^{B_{u}}\mathbbm{1}(\textup{max}(\hat{q}^{u}_{i}) \geq \tau))H(\hat{q}^{u}_{i}, Q^{u}_{i})
\label{eq:fixmatch}
\end{equation} 
\normalsize
 where $\mathbbm{1}( \cdot)$ is the indicator function.
 
Unlike the fixed threshold approach, which uses a single threshold for all classes, the flexible threshold approach in FlexMatch~\cite{zhang2021flexmatch} sets and updates different thresholds for each class. In Eq.\ref{eq:fixmatch}, $\tau$ is replaced with $T=[ \tau_{1}, \cdots, \tau_{c}]$, where $c$ is the number of classes. The flexible threshold for each class is represented by $\tau(\textup{argmax}(\hat{q}^{u}_{i}))$, where $\textup{argmax}(\hat{q}^{u}_{i})$ identifies the class with the highest predicted probability for a given unlabeled sample. The loss function for unlabeled data, as shown in Eq.\ref{eq:flexmatch}, is then defined as:

\vspace{-5pt}
\small
\begin{equation}
L_{u}=\frac{1}{B_{u}}\sum_{i}^{B_{u}}\mathbbm{1}(\textup{max}(\hat{q}^{u}_{i}) \geq T(\textup{argmax}(\hat{q}^{u}_{i})))H(\hat{q}^{u}_{i}, Q^{u}_{i}).
\label{eq:flexmatch}
\end{equation}
\normalsize

\subsection{Model Pipeline}\label{sec:model_pipeline}
Our model follows the state-of-the-art semi-supervised learning framework for video action recognition, SVFormer~\cite{xing2023svformer}. This framework utilizes the consistency loss based on FixMatch, along with augmentation methods for video modality, namely TubeToken Mix (TTMix) and Temporal Warping Augmentation (TWAug). 

Let two unlabeled video clips as $v^{u}_{a}, v^{u}_{b} \in \mathbbm{R}^{H \times W \times C \times T}$ where $(H, W)$ is the resolution of the image frame, $C$ is the number of the channels and $T$ is the number of the frames in the video clip, and let two embedded vectors $Emb_{vid}(v^{u}_{a}), Emb_{vid}(v^{u}_{b}) \in \mathbbm{R}^{N \times (P^{2} \cdot C) \times T}$ where $P$ is the non-overlapping patch size and $N=HW/P^{2}$ is the number of the embedding patches. For the mixing of two embedded video clips, token-level mask $\mathbf{M} \in \{0, 1\}^{(P^{2} \cdot C) \times T}$ is utilized as Eq.~\ref{eq:mixup}.

\small
\begin{equation}
\begin{aligned}
    e^{u}_{mix} = &Emb_{vid}(\mathit{A}_{vid}(v^{u}_{a})) \odot \mathbf{M} \\ & + Emb_{vid}(\mathit{A}_{vid}(v^{u}_{b})) \odot (1-\mathbf{M})
\label{eq:mixup}
\end{aligned}
\end{equation} \normalsize
where $\odot$ is element-wise multiplication.

At this point, SVFormer utilizes tube-style masking, which shares the same mask pattern through all the frames and has consistency in the temporal axis, rather than frame token masking and random token masking. In the case of TWAug, it is utilized for stretching one frame to various temporal lengths. For example, given the number of $T$ frames, they select a few frames and pad with chosen frames at the other frame positions that are not selected. By this TWAug, they argue that the model can learn temporal dynamics flexibly.

Although most commonly encountered videos contain sequences of frames and corresponding audio information, many studies pay less attention to audio information. Inspired by this respect, we expand the SVFormer framework to video-audio modalities. 

Let unlabeled audio log mel-filterbank feature data as $a^{u}_{a}, a^{u}_{b} \in \mathbbm{R}^{M \times L}$ from corresponding video clips $v^{u}_{a}, v^{u}_{b}$, respectively, where $L$ is time-step and $M$ is range of the mel-frequency. For the augmentation, we use SpecAugment~\cite{park2019specaugment} as strong-augmentation $\mathit{A}_{aud}(\cdot)$. Through the Eq.~\ref{eq:aud_mixup}, mixed audio is conducted.

\small
\begin{equation}
    a^{u}_{mix}=\lambda \cdot \mathit{A}_{aud}(a^{u}_{a}) + (1-\lambda) \cdot \mathit{A}_{aud}(a^{u}_{b}) 
\label{eq:aud_mixup}
\end{equation}
\normalsize
where $\lambda$ is the hyper-parameter that follows the beta-distribution used for the mixup.

The following discussion focuses on the process of learning utilizing video and audio data. Initially, a [cls] token is concatenated to the previously obtained embedding, and this combined embedding is then fed into the video encoder $E_{vid}$. As a result, the output $z_{vid}$ is calculated. Similarly, for audio, after embedding the audio, it is input into the audio encoder $E_{aud}$, resulting in the output $z_{aud}$. This calculation is applied in the same way to video clips $v_{a}^{u}, v_{b}^{u}$ and their corresponding audio $a^{u}_{a}, a^{u}_{b}$ as well as to the mixed video and audio obtained from Eq.~\ref{eq:mixup} and \ref{eq:aud_mixup}. This process can be verified through the following equations.

\begin{figure*}
    \centering
    \includegraphics[trim={0cm 1.45cm 2cm 0cm},width=\textwidth, clip]{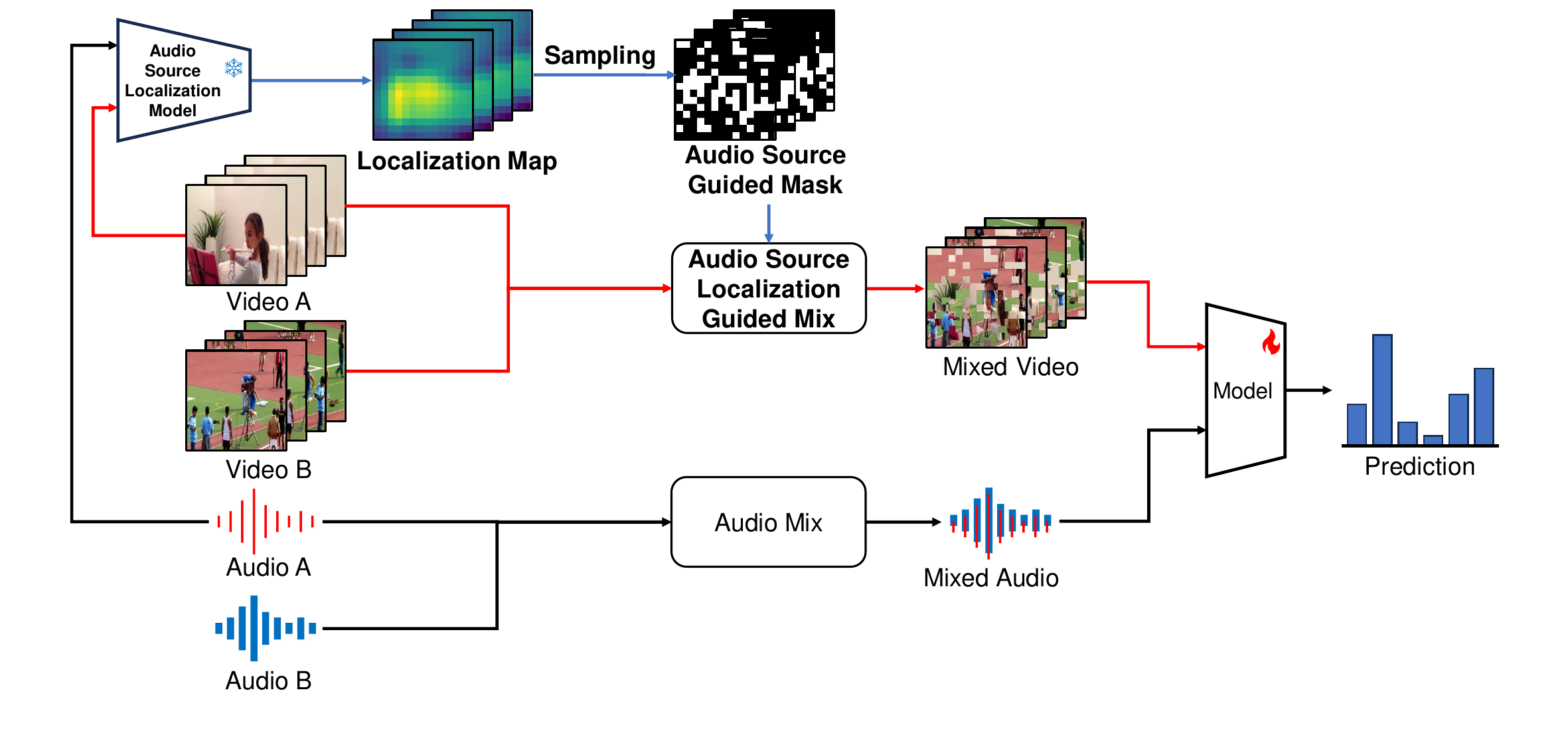}
    \caption{{\bf Overview of our audio source localization-guided mixup framework.} The framework performs the audio source localization-guided mixup on video clips A and B. For generating the audio source localization map, video clip A is utilized. The video and audio from clip A are processed through an audio source localization model to produce the localization maps. This generated map is then used as the weight for performing multinomial sampling without replacement, creating an audio source localization-guided mask. Considering the audio information, this mask guides semantically important regions in video A. Consequently, our proposed audio source localization-guided mixup allows consideration of the interrelation between video and audio modalities sharing the same video clip. For audio A and B, log mel-filterbank coefficients are transformed and interpolated at the pixel level. The resulting mixed video and audio are then used as input for prediction.\vspace{-5pt}
}
    \label{fig:masking}
\end{figure*}
\small
\begin{align}
    z_{vid}&=E_{vid}([\textup{CLS}_{vid}, e_{mix}^u]) \label{eq:logit_video} \\ 
    z_{aud}&=E_{aud}([\textup{CLS}_{aud}, Emb_{aud}(a_{mix}^u)]) \label{eq:logit_audio}\\
    \hat{y}&=\textup{FUSION}(z_{vid}^{0}, z_{aud}^{0}) \label{eq:fusion}
\end{align}
\normalsize
where $z_{vid}^{0}, z_{aud}^{0}$ means the first token of $z_{vid}, z_{aud}$, and $\textup{FUSION}(\cdot)$ means a fusion model consisting of transformer encoder layers.

In the same manner, by forwarding $\alpha_{vid}(v_{a}^{u}), \alpha_{vid}(v_{b}^{u})$ and $a_{a}^{u}, a_{b}^{u}$ into the teacher model, we can calculate $\hat{y}_{a}$ and $\hat{y}_{b}$. Then, using the calculated $\hat{y}_{a}$ and $\hat{y}_{b}$ along with $\lambda$ from Eq.~\ref{eq:aud_mixup}, we compute the pseudo label $\bar{y}_{mix}$. This is used to optimize the model through the consistency loss as described in Eq. \ref{eq:mix_consistency}.

\vspace{-15pt}
\small
\begin{equation}
    L_{mix}=\frac{1}{B_{u}}\sum^{B_{u}}\mathbbm{1}(\textup{max}(\bar{y}_{mix}) \ge \tau) (\bar{y}_{mix} - \hat{y}_{mix})^{2}
    \label{eq:mix_consistency}
\end{equation}
\normalsize

\subsection{Audio Source Localization-guided Mixup}
However, we observe that video and audio are augmented only within their respective modalities in the framework in Section~\ref{sec:model_pipeline}. This approach does not consider the interrelation between the visual and audio information, even though they share the same video clip. Therefore, we propose the novel audio source localization-guided mixup. If the mask is generated through sampling from the audio source localization map and provided as the guide, the relationship between video and audio can be considered when performing the mixup.

We introduce the audio source localization model that is composed of a visual encoder $ E_{vid}^{as}(\cdot) $ and an audio encoder $ E_{aud}^{as}(\cdot) $. For $v^{u}$ and $a^{u}$ sharing a video clip, $ \mathit{A}_{vid}(v^{u}) $ is feed into $ E_{vid}^{as}(\cdot) $, and $\mathit{A}_{aud}(a^{u})$ is similarly feed into $ E_{aud}^{as}(\cdot) $.

As a result, the visual feature $ f_{vid} $ and the audio feature $ f_{aud} $ are calculated. To compute their respective attention maps $ f_{vid}^{attn} $ and $ f_{aud}^{attn} $, we perform matrix multiplication on the transposed features as shown in Eq.~\ref{eq:attention}. Subsequently, we generate the final audio source localization map $ \textup{MAP} $ using dot product operations as outlined in Eq.~\ref{eq:map}.

\begin{figure}
    \centering
    \includegraphics[trim={0cm 10.2cm 24.4cm 0cm},width=\columnwidth, clip]{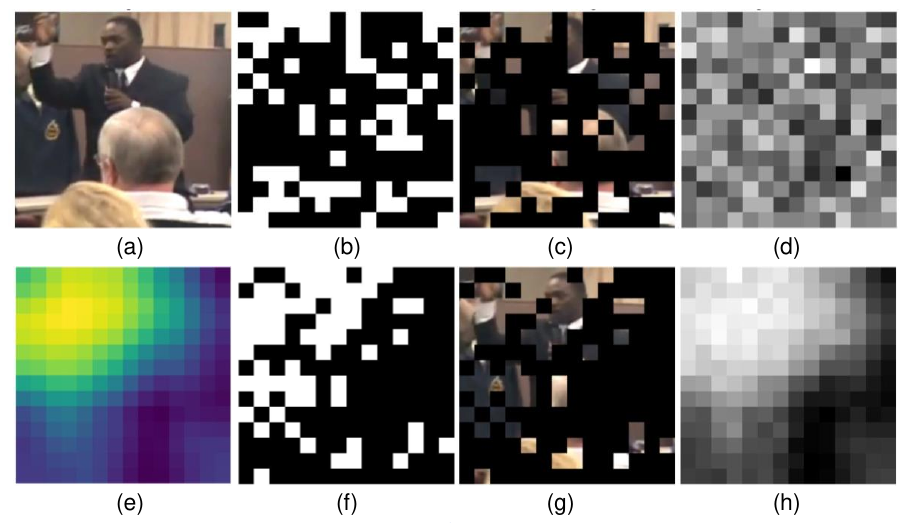}
    \caption{{\bf Visualization of the TubeToken mask and our proposed Audio Source Localization-guided Mask.} 
    \textcolor{\mycolor}{
    When an original image (a) is given, the TubeToken masking creates a random pattern mask (b), resulting in a masked image (c) for SVFormer input. Sampling this mask 255 times yields an average TubeToken Mask (d). Our method uses an audio source localization-guided mask. Starting with the original image (a), a localization map (e) is generated and used as a weight for sampling, creating the guided mask (f). Applying this mask to the original image results in (g). Sampling this mask 255 times produces (h), highlighting audio source areas and allowing for visual-audio modality mixup. \vspace{-10pt}
    }
    }
    \label{fig:visualization}
\end{figure}

\small
\begin{align}
    f_{vid}^{attn} = (f_{vid})(f_{vid})^{T}, \quad &f_{aud}^{attn}= (f_{aud})(f_{aud})^{T} \label{eq:attention} \\
    \textup{MAP} = &(f_{vid}^{attn})(f_{aud}^{attn})^T \label{eq:map}
\end{align}
\normalsize
Building upon this, we can construct the novel mask $ \textup{M}_{as} $ to replace the token-level mask $ \textup{M} $ utilized in Eq.~\ref{eq:mixup}. We interpolate $ \textup{MAP} $ to align with the dimensions of $ Emb_{vid}(v^{u}) $, resulting in an interpolated map denoted as $ \textup{MAP}' $. By normalizing $ \textup{MAP}' $ with min-max normalization, we provide a basis for probabilistically selecting locations likely to contain the audio source. Employing multinomial distribution probabilities, we sample without replacement using $\textup{MAP}'$ as weights. This process generates the new mask $\textup{M}_{as}$, allowing us to retain $(\lambda \cdot N)$ tokens of $Emb{vid}(v^{u})$ for use.
We visualize the differences between our proposed audio source localization-guided mask and the TubeToken mask in Fig.~\ref{fig:visualization}.

\subsection{Visual-Audio Contrastive Learning}
To align visual and audio information, we employ a visual-audio contrastive learning loss. This approach utilizes the embeddings $z^{0}_{vid}$ and $z^{0}_{aud}$, obtained from Eq.~\ref{eq:logit_video} and~\ref{eq:logit_audio}, respectively. Considering labeled, unlabeled, and mixed data, we define the video embeddings matrix $Z_{vid}^{l}, Z_{vid}^{u}, Z_{vid}^{m}$ in Eq.~\ref{eq:concatenated_vid}, where $i \in [B_{l}]$ and $j \in [B_{u}]$. A similar definition applies to the audio embeddings matrix $Z_{aud}$.

First, we perform L2 normalization on $Z_{vid}$ and $Z_{aud}$. The similarity between video and audio is then computed using matrix multiplication as shown in Eq.~\ref{eq:mm_sim}, resulting in the visual-audio similarity matrix $S$.

\small
\begin{equation}
    Z_{vid} = [z^{l_{0, 1}}_{vid}, \cdots, z^{l_{0, i}}_{vid}~~;~~ z^{u_{0, 1}}_{vid}, \cdots, z^{u_{0, j}}_{vid}~~;~~ z^{m_{0, 1}}_{vid}, \cdots, z^{m_{0, j}}_{vid}] 
    \label{eq:concatenated_vid}
\end{equation}
\begin{equation}
    Z_{vid}' = \frac{Z_{vid}}{\|Z_{vid}\|_2}, \quad Z_{aud}' = \frac{Z_{aud}}{\|Z_{aud}\|_2} 
\label{eq:l2_norm}
\end{equation}

\begin{equation}
    S = \frac{1}{0.05} \cdot Z_{aud}'(Z_{vid}')^{T} 
\label{eq:mm_sim}
\end{equation}
\normalsize
Subsequently, we compute the visual-audio contrastive loss $L_c$ using the noise-contrastive estimation loss applied to $S$.

\small
\begin{align}
    \text{NCE}_1 &= -\frac{1}{K} \sum_{i=1}^{K} \log \left(\frac{e^{S_{ii}}}{\sum_{j=1}^{K} e^{S_{ij}}} \right) ,~\\
    \text{NCE}_2 &= -\frac{1}{K} \sum_{i=1}^{K} \log \left( \frac{e^{S_{ii}}}{\sum_{j=1}^{K} e^{S_{ji}}} \right)\\
    L_{c} &= \frac{\text{NCE}_1 + \text{NCE}_2}{2} 
    \label{eq:contrastive}
\end{align}
\normalsize
Here, $K$ denotes the number of elements in $Z_{vid}$ and $Z_{aud}$.

\subsection{Training Objective}
Our proposed model requires four main types of loss functions for training. These include the supervised learning loss (Eq.~\ref{eq:ce-loss}), the consistency regularization loss for unlabeled data (Eq.~\ref{eq:flexmatch}), and the contrastive loss for aligning visual and audio modalities (Eq.~\ref{eq:contrastive}). Additionally, we utilize a consistency regularization loss for a mixture of unlabeled data, guided by audio source localization mixup (Eq.~\ref{eq:mix_consistency}). The total loss used for training is formulated as follows:

\vspace{-5pt}
\small
\begin{equation}
    L_{total} = L_{s} + \gamma_{1} \cdot L_{u} + \gamma_{2} \cdot L_{mix} + \gamma_{3} \cdot L_{c}
\end{equation}
\normalsize
where $\gamma_{1}, \gamma_{2}$, and $ \gamma_{3}$ are the hyper-parameters for balancing the each loss.

\vspace{-5pt}
\section{Experiment}\label{sec:exp}
\vspace{-5pt}
In this section, we first explain the experimental setting. Next, we analyze the experimental results conducted with different numbers of labeled data on different datasets and discuss ablation studies.
\vspace{-5pt}

\subsection{Experimental Setting}
\paragraph{\bf{Datasets}}
We conduct experiments using three datasets: UCF-51~\cite{soomro2012ucf101}, Kinetics-400~\cite{kay2017kinetics}, and VGGSound~\cite{chen2020vggsound}. UCF-51 is a refined subset of the UCF-101 dataset, comprising 51 classes, including audio, consisting of 4.9K training and 1.4K test samples. Kinetics-400 covers 400 categories, each with approximately 10-second-long videos, amounting to 240K training and 20K validation samples. For this dataset, samples without audio are filtered out. VGGSound is composed of around 200K videos, each about 10 seconds in length, and provides annotations for 309 categories. After excluding unavailable videos, we have 183,730 training samples and 15,446 test samples. Before the experiments, the UCF-51, Kinetics-400, and VGGSound datasets are split to ensure each class contains 1 and 5 labeled data samples, respectively. Using one labeled data sample per class is a more extremely challenging experimental condition, which has also been used in semi-supervised image classification studies~\cite{wang2022freematch, lucas2022barely}. Additionally, using five labeled data samples per class is similar to other studies~\cite{xing2023svformer, xu2022cross, xiong2021multiview} that take a 1\% labeled samples setting, which uniformly uses only six labeled samples per class, despite the dataset being imbalanced.
\vspace{-7pt}

\begin{table*}[ht]
\resizebox{\textwidth}{!}{
\begin{tabular}{lccccccccccc}
\toprule
\multirow{2}{*}{Model}    & \multirow{2}{*}{Backbone} & \multirow{2}{*}{Input} & \multirow{2}{*}{Epoch} & \multirow{2}{*}{\begin{tabular}[c]{@{}c@{}}Threshold\\ Type\end{tabular}} & \multirow{2}{*}{\begin{tabular}[c]{@{}c@{}}Mask\\ Type\end{tabular}} & \multicolumn{2}{c}{UCF-51}      & \multicolumn{2}{c}{Kinetic400}  & \multicolumn{2}{c}{VGGSound}    \\ \cline{7-12} 
                          &                           &                        &                        &                                                                           &                                                                      & 51             & 255            & 400            & 2000           & 309            & 1545           \\ \hline
TCL~\cite{singh2021semi}                       & TSM-Res18               & V                    & 400                    & Fix                                                                       & -                                                                    & 9.16          & 25.93          & 1.18           & 2.26           & 0.43           & 0.66           \\
LTG~\cite{xiao2022learning}                       & 3D-Res18               & V, G                    & 180                    & Fix                                                                       & -                                                                    & 16.51          & 49.74          & 0.46           & 3.39           & 0.37           & 0.45           \\
AvCLR~\cite{assefa2023audio}                     & 3D-Res18               & V, A                   & 800                    & Fix                                                                       & -                                                                    & -              & 50.1           & -              & -              & -              & -              \\ \hline
\multirow{2}{*}{SVFormer~\cite{xing2023svformer}} & ViT-B                     & V                      & 50                     & Fix                                                                       & TubeToken                                                            & 60.75          & 87.71    & 16.01          & 46.93          & 14.87          & 37.70    \\
                          & ViT-B                     & V                      & 50                     & Flex                                                                      & TubeToken                                                            & 63.58    & 86.63          & 16.69    & 47.62    & 15.98    & 36.72          \\ \hline
Ours                      & ViT-B                     & V, A                   & 50                     & Flex                                                                      & ASL-Guided                                                           & \textbf{72.58} & \textbf{89.09} & \textbf{19.12} & \textbf{48.64} & \textbf{17.49} & \textbf{38.00} \\ \bottomrule
\end{tabular}}
\caption{ 
{\bf Top-1 accuracy comparisons with state-of-the-art methods on UCF-51, Kinetics-400, and VGGSound.} Note that TCL, LTG, and AvCLR are CNN-based, while SVFormer and ours use Transformer architectures. `V', `A', and `G' represent video, audio, and temporal gradient modalities. `Fix' refers to the fixed threshold from FixMatch~\cite{sohn2020fixmatch}, and `Flex' to the flexible threshold from FlexMatch~\cite{zhang2021flexmatch}. `TubeToken' is SVFormer's mask strategy, and `ASL-Guided' is our proposed Audio Source Localization-Guided method. The numbers below each dataset show the amount of labeled data used in training.}
\label{tb:main}
\end{table*}

\vspace{-5pt}
\paragraph{\bf{Baselines}}
We use the prior studies LTG~\cite{xiao2022learning}, SVFormer~\cite{xing2023svformer}, and the state-of-the-art audio-visual semi-supervised action recognition, AvCLR~\cite{assefa2023audio}, as our baselines. However, in the case of AvCLR, the implementation codes are not public, so we report the results of their study. Additionally, we employ a model that modifies the fixed threshold used in the original SVFormer to a flexible threshold. The majority of the hyper-parameters are set up identically to those in SVFormer. 

\vspace{-10pt}
\paragraph{\bf{Evaluation}} 
\textcolor{\mycolor}{
We utilize top-1 accuracy as the metric. Given the potential unfairness of comparing CNN-based methods directly with Transformer-based methods, it is worth noting that our comparison is made directly with SVFormer, the state-of-the-art semi-supervised video action recognition model, instead of directly comparing with CNN-based.
}
\vspace{-12pt}

\subsection{Implementation Details}
For training, we follow the SVFormer settings. All experiments are conducted with 2 $\times$ A100 80GB GPUs. We utilize the FNAC~\cite{sun2023learning} as the audio source localization model. FNAC is a two-stream network comprising two ResNet-18 models, serving as the visual and audio encoders, respectively. During the training phase, we use a version of FNAC pre-trained on the Flickr-1K dataset~\cite{plummer2015flickr30k} without object-guided localization. This model is kept frozen and is not further trained during our training phase. 
We set the ratio between labeled and unlabeled samples as 1:5 in the mini-batch following SVFormer's setting. Also, we use an SGD optimizer for training, with a momentum of 0.9 and a weight decay of 0.001. 
The exponential moving average from the student model serves as the teacher model in both SVFormer and ours.

The values of $\gamma_{1}$ and $\gamma_{2}$ are set to 2, while for $\gamma_{3}$, we select the optimal value from \{0.1, 0.2, 0.3\}. Additionally, for the masking ratio $\lambda$, we sample from the beta distribution Beta($\alpha_{1}$, $\alpha_{2}$), setting $\alpha_{1}$ to 5 and $\alpha_{2}$ to 10. This is because, guided by audio source localization, it is acceptable to take a smaller ratio of the video for creating the localization map due to the selection of significant regions. During the testing phase, the entire video is uniformly divided into five parts, and we perform cropping three times to cover most areas of the video clip, each with a size of $224 \times 224$. For the final testing prediction, we averaged a total of 15 predictions obtained by performing the aforementioned three crops five times per video clip.

\begin{table}
\resizebox{\columnwidth}{!}{
\centering
\begin{tabular}{lccccc}
\toprule
Input Type & Mask Type                 & Contrastive               & Trainable Params. & GFLOPs  & Accuracy          \\ \hline
V          & TubeToken                 &                           & 121M              & 761.144 & 63.58             \\ \hline
V,A        & TubeToken                 &                           & 254M              & 805.167 & 66.72             \\ 
V,A        & TubeToken                 & \checkmark                & 254M              & 805.167 & 68.52             \\ \hline
V,A        & Proposed                  &                           & 254M              & 820.998 & \underline{71.55} \\
V,A        & Proposed                  & \checkmark                & 254M              & 820.998 & \textbf{72.58}    \\ \bottomrule
\end{tabular}}
\caption{{\bf Ablation study about using audio source localization-guided mask and contrastive learning.} This experiment is conducted on the UCF-51 dataset using one labeled sample per class. It aims to understand the effectiveness of our proposed ASL mask and audio-visual contrastive learning. \vspace{-7pt}}
\label{tb:mask_contrastive}
\end{table}

\subsection{Main Results}
The main experimental results on UCF-51, Kinetics-400, and VGGSound datasets can be found in Table~\ref{tb:main}. Compared to previous methods, our proposed method demonstrates superior performance. Specifically, on the UCF-51 dataset, using only one labeled sample per class, we achieve an absolute performance improvement of 9.00\% accuracy and also a relative accuracy improvement of 14.1\% over the SVFormer, which only uses video modality with flexible thresholding. 
Moreover, on Kinetics-400, our model outperforms SVFormer by 2.43\% and 15.56\%, and on VGGSound, we achieve  1.51\% and 9.45\% higher performances in absolute and relative accuracy improvement, respectively.
Even though the gains become smaller when considering performance with five samples per class, which is easier, our approach still outperforms baselines and improves performance. These results demonstrate the effectiveness of our proposed audio source localization-guided mixup method in considering the intermodal relation between video and audio modalities.

Our method, which leverages the visual-audio modality in a semi-supervised learning approach and considers the relationship between video and audio, outperforms the state-of-the-art CNN-based AvCLR by 38.99\% on UCF-51. However, due to the unavailability of AvCLR's code, we are limited to comparing it against its published results only. This limitation may raise concerns about whether the performance difference is solely due to the backbone architecture (CNN vs. Transformer). To address these concerns and demonstrate that the effectiveness of our method goes beyond architectural differences, we provide additional experiments in Section~\ref{sec:ablaction}.

\subsection{Ablation Studies}\label{sec:ablaction}
To understand the impacts of the proposed methods, we conduct the ablation studies on UCF-51.

\vspace{-10pt}
\paragraph{\bf{Analysis of Mask Type and Contrastive Learning}}
To evaluate the effectiveness of our proposed methodologies, we conduct an ablation study focusing on the audio source localization-guided mask (ASL mask) and visual-audio contrastive learning, and the result is in Table~\ref{tb:mask_contrastive}. Initially, we observe that integrating both video and audio modalities enhances performance by a margin of 3.14\%. Furthermore, when applying visual-audio contrastive learning, we note an additional performance increase of 1.80\%.
Even though the integration of an additional audio encoder and fusion module increases the number of trainable parameters to utilize the audio modality, it significantly boosts overall performance. The empirical evidence supports that the utilization of audio modality contributes positively.

\begin{figure}
\centering
     \includegraphics[trim={0cm 11.6cm 27.6cm 0cm}, width=1\linewidth, clip]{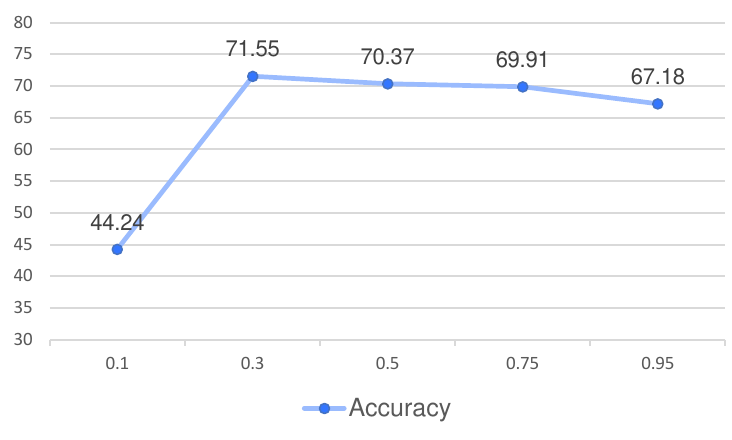}
    \caption{{\bf Impact of threshold ($\tau$).} This figure presents the results of experiments conducted to explore the performance impact of the hyper-parameter $\tau$. It specifically focuses on the changes in performance with varying thresholds of $\tau$ during training with only one labeled sample in the UCF-51 dataset.\vspace{-10pt}}
    \label{fig:threshold}
\end{figure}

A noteworthy outcome of our study is the performance improvement of 4.83\% achieved by employing the ASL mask over the TubeToken mask. This improvement becomes even more significant when the ASL mask is combined with contrastive learning, achieving an accuracy of 72.58\%. 
These findings demonstrate the effectiveness of our audio source localization-guided mixup in considering the inter-modal relation between video and audio modalities. At the same time, visual-audio contrastive learning contributes to further performance enhancements.

The additional computations obtained by adding audio modality and applying the proposed mask strategy are small compared to existing video computations in a single training iteration.
Considering the substantial performance gains, the additional computational load is manageable.

\paragraph{\bf{Threshold for Pseudo Label}}
We conduct an ablation study to assess the impact of the threshold $\tau$ for generating pseudo labels in our proposed audio source localization guided-mixup method. In this study, the predicted probability calculated by the teacher model is used as a pseudo label, and we explore how different $\tau$ values affect this process. 

We observed that a lower threshold of 0.1 generally results in most predicted probabilities being used as pseudo labels. However, this can lead to performance degradation due to the accumulation of bias from low-confidence pseudo-labels. 

Moreover, ours of the flexible thresholding method proposed in FlexMatch~\cite{zhang2021flexmatch} allows us to maintain superior performance compared to SVFormer, which also incorporates audio modality, even as $\tau$ increases. Among various $\tau$ values, we identified 0.3 as the optimal value and set it for our main experiments and other ablation studies. The performance variations according to different $\tau$ values can be seen in Fig.~\ref{fig:threshold}.

\begin{figure}
\centering
    \includegraphics[trim={0cm 11.6cm 27.6cm 0cm},width=\columnwidth, clip]
    {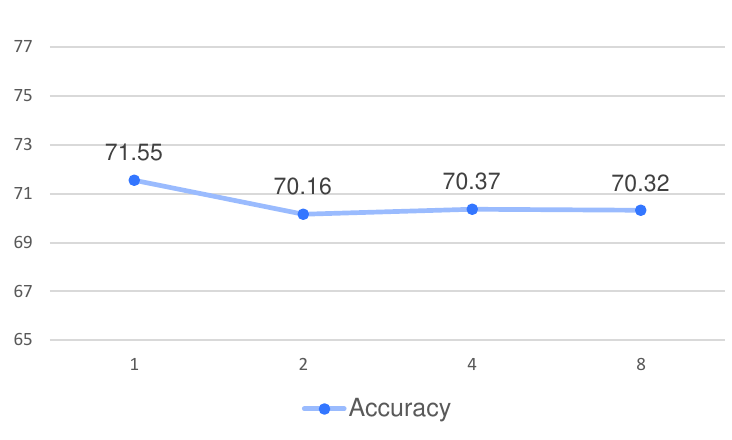}
         \caption{{\bf Ablation study on the effect of varying frame counts (1, 2, 4, and 8 frames) on the audio source localization map.} This study evaluates the impact of averaging the localization maps over different numbers of frames before using them in the sampling process.\vspace{-10pt}
         }\label{fig:num_avg_mask}
\end{figure}

\paragraph{\bf{Mask Sampling per Varying Frames}}
Inspired by the applications of tube-shaped masks in the video domain~\cite{xing2023svformer, tong2022videomae}, which utilize consistent tokens across time to prevent information leakage from adjacent frames, we conduct an ablation study on the creation of audio source localization masks. 
This study involves averaging maps generated from adjacent frames and using these averages for sampling to create masks. Our experiments are designed to test averages over {1, 2, 4, 8} frames. The results of this experiment can be seen in Fig.~\ref{fig:num_avg_mask}.

From our findings, we observe that contrary to the TubeToken mask approach, generating masks based on each individual frame achieves the best performance for our proposed audio source localization-guided masks. This can be attributed to the fact that while audio source localization maps are generated for each frame based on visual and audio information, the visual information varies from frame to frame. 
Therefore, when using the average of audio source localization maps generated from different frames, it can prevent information leakage from adjacent frames during token-level mixup, similar to the TubeToken approach. This method might be effective for static videos with minimal audio source location changes. However, for videos with rapid changes, this averaging approach uses a mean of different audio source locations, which may not accurately represent the true location in dynamic scenes. Consequently, this can lead to significant information loss and, ultimately, a decrease in performance.

\section{Conclusion}\label{sec:conclusion}
\textcolor{\mycolor}{
This paper introduces a transformer-based semi-supervised multimodal video action recognition approach, expanding the previous visual framework to include visual-audio information from video clips. To address the limitations of existing augmentation methods that only consider individual modalities, we propose the audio source localization-guided mixup, which considers the interrelation between visual and audio information. Our method demonstrates superior performance over state-of-the-art methods on the UCF-51, Kinetics-400, and VGGSound datasets.
While our approach relies on the audio source localization method, we anticipate that our performance will further improve as audio source localization techniques advance. Incorporating audio modalities into SVFormer~\cite{xing2023svformer} requires an additional encoder and fusion module, which increases computational costs. 
Nevertheless, our method enhances performance and explores the relationships between visual and audio modalities, demonstrating that a comprehensive understanding of these relationships can significantly boost performance.
}



{\small
\bibliographystyle{ieee_fullname}
\bibliography{main}
}

\end{document}